AIRES 2021 Research Conference: *Democratization* of AI

# Good AI For The Present of Humanity Democratizing AI Governance

*Nicholas Kluge Corrêa₁, Nythamar de Oliveria₁*
*₁Department of Philosophy, Pontifical Catholic University of Rio Grande do Sul, Brazil*

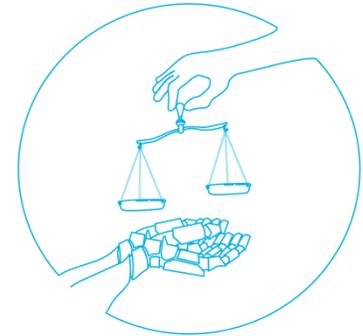

## Abstract

What do Cyberpunk and AI Ethics have to do with each other? Cyberpunk is a sub-genre of science fiction that explores the post-human relationships between human experience and technology. One similarity between AI Ethics and Cyberpunk literature is that both seek to explore future social and ethical problems that our technological advances may bring upon society. In recent years, an increasing number of ethical matters involving AI have been pointed out and debated, and several ethical principles and guides have been suggested as governance policies for the tech industry. However, would this be the role of AI Ethics? To serve as a soft and ambiguous version of the law? We would like to advocate in this article for a more Cyberpunk way of doing AI Ethics, with a more democratic way of governance. In this study, we will seek to expose some of the deficits of the underlying power structures of the AI industry, and suggest that AI governance be subject to public opinion, so that 'good AI' can become 'good AI for all.'

**History**
Received **1 April 2021**
Accepted **14 July 2021**
Published **16 July 2021**

**Contact**

**Nicholas Kluge Corrêa**
Ph.D Student at The Pontifical Catholic University of Rio Grande do Sul - Department of Philosophy, Brazil
(nicholas.correa@acad.pucrs.br)
+55(51)981706812

**Nythamar de Oliveira**
Full Professor at The Pontifical Catholic University of Rio Grande do Sul Department of Philosophy, Brazil
(nythamar@yahoo.com)

**Acknowledgements**

**None**

**Disclosure of Funding**

None

## I.    Introduction

*For a More 'Cyberpunk' Way of Conducting AI Ethics*

With the ever-growing advancements in Artificial Intelligence (AI), autonomous systems are increasingly becoming a part of our society, with novel technologies such as robotics, nanotechnology, genetics, and artificial intelligence, promising to transform our world and the way we live (Mulhall, 2002). At the present moment, the most accessible and massively used technology in our society, of those mentioned above, is AI. Given the size and complexity that our society has grown, human beings alone are not able to cope with the demands of processes that are vital to our civilization, and we increasingly rely on the help of intelligent automation.

We realize that, perhaps without many controversies, our current society cannot exist in its present form without the help of such technologies. Samuel Butler (1863), in his work 'Darwin Among the Machines', questioned our 'quiescent bondage' to technology. Butler argued that one day we would reach the point where society would no longer be able to separate itself from its technological creations because it would be equivalent to the suicide of the status quo. In Butler's words: '[…] this at once proves that the mischief is already done, that our servitude has commenced […].'

In the end, whether all the technological modernization we experience will result in a future good for all humanity is still a question with no answer. And many believe that this is an answer worth pursuing sooner rather than later. We agree thus with Feenberg (2017) in that the critique of hubris is the basis for an ethic and a politics of technology, as we renounce the illusion of godlike power to master nature and bend it to our will through technology, given our human finitude. In effect, we need a realistic, balanced view of both technology and ethics without demonizing or idolizing the former and avoiding normativism and dogmatism when dealing with the latter.

In sociological and literary terms, contemporary critical theory, with its origins in sociology and literary criticism, proposes to conduct a reflexive and critical assessment of society and culture to reveal and challenge deficits in their underlying power structures. We propose that there is a fruitful relationship between the criticism made by contemporary critical theorists, like Craig Calhoun (1995), Paul Virilio (1997), Hartmut Rosa (2010), and Andrew Feenberg (2017) that can help ethics be more 'what it was meant to be.'

We also would like to point out that while contemporary critical theory focuses on the present, another possible form of criticism involves extrapolating the future. Cyberpunk, a subgenre of science fiction, seeks to show how our technological advances can lead our society to dystopian outcomes, and the ethical and social problems which may be ahead. Authors such as Philip K. Dick (Do Androids Dream of Electric Sheep?), John Brunner (Stand on Zanzibar), William Gibson (Neuromancer), surrounded by the technological innovations of the '80s and '90s (internet, AI, robotics, virtual reality, genetics), gave rise to a form of literature aimed to criticize certain aspects of the postmodern condition.

Fredric Jameson defines cyberpunk as '[...] the supreme literal expression, if not of postmodernism, then of late capitalism itself' (Jameson, 1991, p. 417). Similar to Jameson, Jean Baudrillard (1994) proposed that given the rapid pace of social and cultural transformation we are experiencing, sociological studies are increasingly approaching what we call science fiction, where we progressively need to anticipate social change while it is happening.

When it comes to ethical and philosophical debates, what

# AI Ethics Journal

we see today is a kind of 'soft' response to the postmodern critique of cyberpunk, that is: how can we avoid the blind march into the dystopian future? How can we avoid the emergence of increasingly authoritarian and technocratic states? In this context, the premise for security issues involving our technological advances is established on an idea of a negative utopia. In the words of Robert Tally:

> *"First of all, the utopian impulse must be negative: identify the problem or problems that must be corrected. Far from presenting an idyllic, happy and fulfilled world, utopias should initially present the root causes of society's ills [...] to act as a criticism of the existing system"* (Tally, 2009, p. 11).

We can thus say that the critique proposed by some critical theorists, some postmodern sociologists and philosophers, and Cyberpunk is a manifestation of the negative utopian impulse. But do we see this spirit of critique in the current debate of AI Ethics? In our opinion, very little. What we see is a great number of Ethical Guidelines being proposed to regulate the tech industry (Russell et al., 2015; Amodei et al., 2016; Boddington, 2017; Goldsmith & Burton, 2017; Greene et al., 2019). But of course, *if* the industry chooses to follow them. It's not as if they were laws.

Would all these published ethical guidelines have any real normative power over the AI industry? Like Ryan Calo (2017), we also think that ethical guidelines end up serving more as a marketing strategy than a real effort to regulate the tech industry:

> *Several efforts are underway, within the industry, academia, and several NGO's, to resolve the ethics of AI. But these efforts probably cannot replace policy-making* (Calo, 2017, p. 407).

To support this claim, we might evoke a controlled study conducted by McNamara et al. (2018). The sole purpose of this study was to investigate whether ethical guidelines have a normative effect on the decision-making of software developers. In their research, the authors evaluated 63 software engineering students and 105 professional software developers, analyzing whether the ethical guidelines from the Association for Computing Machines[1] (ACM) would have any influence on moral dilemmas related to software production. The results: 'Despite our stated goal, we found no evidence that the code of ethics of the ACM influences ethical decision-making.' Maybe this study shows that there is a hole in the academic education of software developers, like applied ethics. Or, perhaps we are just inefficiently using Ethics.

Several studies support the idea that ethical guidelines have little to no effect on decision-making in many different professional fields (Brief et al., 1996; Cleek & Leonard, 1998; Lere & Guamnitz, 2003; Osborn et al., 2009). And this idea resonates with several criticisms raised against the current state of AI Ethics. Jobin et al. (2019, p. 389):

> *'Private sector involvement in the field of AI ethics has been questioned for potentially using soft policies as a way to turn a social problem into something technical or to completely avoid regulation.'*

Hagendorff (2020, p. 99):

> *'AI ethics - or ethics in general - have no mechanisms to reinforce its normative claims.'*

Rességuier and Rodrigues (2020, p. 1):

> *'Ethics have great powerful teeth. Unfortunately, we are barely using them in AI ethics - no wonder then that AI ethics is called toothless.'*

---

[1] ACM Code of Ethics and Professional Conduct. http://www.acm.org/binaries/content/assets/membership/images2/fac-stu-poster-code.pdf.

And last, Mittelstadt:

> *Statements reliant on vague normative concepts hide points of political and ethical conflict. 'Fairness', 'dignity', and other such abstract concepts are examples of 'essentially contested concepts'* [...] *At best, this conceptual ambiguity allows for the context-sensitive specification of ethical requirements for AI. At worst, it masks fundamental, principled disagreement and drives AI Ethics towards moral relativism. At a minimum, any compromise reached thus far around core principles for AI Ethics does not reflect meaningful consensus on a common practical direction for 'good' AI development and governance* (Mittelstadt, 2019, p. 503).

The point we want to make in this introduction is that the role of ethics is not to be a soft version of the law, even if laws are based on ethical principles. That is not where ethics finds its normative thrust.

Like critical theory and literature, the application of ethics lies in challenging the status quo, seeking its deficits and blind spots. Ethicists concerned with the current state of the AI industry shouldn't only reinforce the repetition of the same concepts already cited by numerous published guidelines. Guidelines that commonly are made by the AI industry, which are (weirdly) self-regulating itself. But we should seek to (re)visit all the issues that are being overlooked. Issues like diversity, representativeness, anti-war policies, equality of income and wealth distribution, the preservation of our socio-ecological system, things that are rarely cited in these Ethical Guidelines.

## II. Safety Issues and AI Ethics
*Technical and Social Problems*

Now, what do these Ethical Guidelines claim after all? Who makes this? In their meta-analysis, Jobin et al. (2019) mapped all the countries responsible for producing the existing ethical guidelines for AI regulation. Their research identified 84 documents containing ethical guidelines for intelligent autonomous systems, that converged around five ethical principles; transparency, justice, non-maleficence, responsibility, and privacy. Hagendorff's (2020) meta-analysis of the main ethical guidelines published in the last five years showed that the main ethical principles cited by them were similar to Jobin et al. (2019) findings, accountability, explainability, and privacy, appearing in almost all guidelines. These principles can be described as follows:

1. *Accountability:* how to make the AI industry accountable for its technologies. For example, in the context of autonomous vehicles, what kind of guarantees and responsibilities should companies developing autonomous vehicles offer to society?

2. *Explainability:* one of the greatest shortcomings in contemporary machine learning systems is that it is difficult to explain the internal process of these types of AI systems, especially when using architectures like deep neural nets.

3. *Privacy:* The abundance of data that we produce daily ensures an inexhaustible source of information for the training of AI systems. However, the use of personal data without consent is one of the main preoccupations found in the literature involving AI Ethics.

Jurić et al. (2020) conducted a similar study, a quantitative bibliographic survey on the recent expansion of AI safety research and its main topics of interest. The common motivation for short and long-term interests in AI safety and AI Ethics is the same: how to make the interaction between humans and AI safe and beneficial? And this is what a lot of the contemporary debate on AI Ethics has delimited itself. Questions like how to make possible advanced AIs operating by reinforcement learning corrigible (Soares et al., 2015; Turner et al., 2020), or how to align the terminal goals of AI systems with our values (Soares, 2016; Russel, 2019), and even how to integrate human society in a post-Singularity era (Chalmers, 2010).

As much as anyone interested in AI safety (with a properly calibrated moral compass), the authors do not want powerfully misaligned AI systems turning their future light-cone into paper clips or anything like that. Nor do we desire any form of hellish dystopian Singularity desolated future, and probably no one does. Not wanting to reduce the importance of Alignment research and all the benefits it may bring to future humanity, we ask, *what about present humanity*? What are we doing to prevent the side-effects of AI and mass automation right now? Who will survive to enjoy the pleasures of aligned AI in the future?

Furthermore, Hagendorff (2020) points out in his meta-analysis that the main principles, Accountability, Explainability, and Privacy, mentioned in the most recent published ethical guidelines have a considerable research effort to ensure aspects such as transparency, legal accountability, and preservation of privacy in the literature. However, of the 22 most relevant published ethical guidelines in the last five years, only nine mention labor rights and technological unemployment, while only two mention the lack of diversity in the AI industry.

Since some of the most underrated problems end up being related to social inclusion and respect for diversity, we think it is fair that we gave them a little bit more room in the current AI ethical debate, so we can see the current humanitarian cost and social risks we are facing, with 'weak' AI being run by a misalign world.

In a practical sense, what we need is regulation, not just suggestions. And the role of AI Ethics should not only be to suggest, but also to criticize. In this spirit, in the following sections, we would like to offer a brief critique of the lack of diversity found in the AI Industry, pointing out the side effects caused by this general state of inequality.

We also believe that regulations should be able to put the AI industry in check, should its technologies promote, for example, crimes against humanity (e. g. autonomous weapons used against civilians). But how can we achieve this?

At the end of this study, we will suggest that 'technological crimes', or 'cyber-crimes', should be defined as violations of universal/international rights (e. g., the International Humanitarian Law or the Universal Declaration of Human Rights). And by doing so, use international agencies, in addition to local (national) agencies, to prosecute actions that violate the wellbeing of civil society. Keeping in mind that the 'Social Good' must always be defined in local terms, but its protection must be a question of universal care.

## III. Who Will Lose Their Job

In the last two centuries, many jobs and forms of occupation have not lasted more than 100 years in our society (for example, telephone operators, typists, public pole lighters). Nowadays, through AI, companies can drastically reduce their need for human labor to lower their costs. However, the adoption of this management policy has two obvious consequences:

1. Wealth accumulation for companies oriented to the development of AI;

2. The unemployed population replaced by automation would find themselves without any source of income.

# AI Ethics Journal

This reality is best summarized by Erik Brynjolfsson[2] in the following quote:

> *It is one of the dirty secrets of the economy: technological progress makes the economy grow and creates wealth, but there is no economic law that says everyone will benefit.*

Frey and Osborne (2013) estimated the probability of automation for 702 occupations in the USA. The result showed that approximately 47% of these occupations will be eliminated by technology over the next 20 years. This estimate can be used in other regions of the world, such as Latin America, which, according to a study published by the International Labor Organization (ILO), breaks a historical record of unemployment[3] in 2020.

In the second quarter of 2020, Brazil registered 12.8 million unemployed people[4], 1.8 million more than at the end of 2019[5]. The pandemic of the new coronavirus has helped accelerate job losses in Call centers, a sector that is already being rapidly automated by chatbots and virtual assistants (Hao, 2020). Even Academics are not safe. In May 2020 more than 90 university professors were fired from the Laureate group, responsible for universities such as Anhembi Morumbi, the FMU University Center, and other universities in Brazil. The fired professionals were all responsible for teaching disciplines in a distance education format. The Laureate group replaced these professionals with 'monitors' and autonomous tools for proofreading (Domenici, 2020).

Although autonomous vehicles are not yet a publicly available technology, their test versions are circulating in several places. It is estimated that by 2021 at least five major automotive companies will have autonomous cars and trucks available for the general public (Maxmen 2018). But what will be the effect of automation on the transportation industry and its workers? How much of the working population will be affected? According to the company Uber[6], more than one million drivers work for the company in Brazil. According to the Brazilian National Agency for Land Transport[7], the country's truck fleet has 1.941 million units. Of this total, 703,000 vehicles are owned by independent truck drivers.

Meanwhile, data regarding the number of delivery workers in Brazil is difficult to obtain. According to Eufrásio and Goulart (2020), the company iFood has registered 170 thousand deliverers in Brazil, and the Rappi platform has 200 thousand deliverers throughout Latin America. Brazil also has the largest number of motorcycle delivery workers in the world, according to Sindimoto - SP[8] (São Paulo State Union of Motorcycle, Cyclist, and Mototaxi Messengers), 2017 Brazil had more than 1.85 million motorcycle delivery workers (representing 30% of the workforce in Brazil). Would our social support network be ready to deal with this demand?

Another preoccupying aspect is the perceptible growth of informal work alternatives (Antunes, 2009), characterized as another humanitarian problem involving labor exploitation. An easy-to-cite example is the emergence of click working. Click work is a type of essential task for training AI systems. Usually, machine learning requires large amounts of labeled data to become proficient in

---

[2] Interview by Rotman, D. 2013. How technology is destroying jobs. MIT Technology Review. https://www.technologyreview.com/2013/06/12/178008/how-technology-is-destroying-jobs/.
[3] https://www.ilo.org/caribbean/newsroom/WCMS_749692/lang--en/index.htm.
[4] https://www.ibge.gov.br/explica/desemprego.php.
[5] https://www.ibge.gov.br/estatisticas/sociais/trabalho/9173-pesquisa-nacional-por-amostra-de-domicilios-continua-trimestral.html?=&t=series-historicas&utm_source=landing&utm_medium=explica&utm_campaign=desemprego.
[6] https://www.uber.com/pt-BR/newsroom/fatos-e-dados-sobre-uber/.
[7] https://agenciabrasil.ebc.com.br/economia/noticia/2019-12/governo-lanca-programa-de-incentivo-caminhoneiros-autonomos#:~:text=Segundo%20dados%20da%20Ag%C3%AAncia%20Nacional,apenas%2026%20mil%20s%C3%A3o%20cooperados.
[8] http://www.sindimotosp.com.br/noticias/noticia146.html.

# AI Ethics Journal

certain tasks. Thus, human individuals are hired to perform tasks such as image classification (Irani, 2016). Companies that hire such a workforce usually do not offer minimum wage, sometimes even charging commissions for each transaction made by the workers. These workers still experience difficulty in receiving payment, obtaining technical assistance, or any other kind of support from the companies they work for (Harris, 2014).

The fact that large AI companies pay pennies for the kind of essential work that makes machines learning efficient and valuable demonstrates a certain indifference on the part of these companies to notions of economic egalitarianism, income equality, and the value of human labor. Without analyzing these new labor relations under the light of the ILO guidelines put the lives of these individuals at risk. According to the ILO Declaration on Social Justice for a Fair Globalization made in 2008[9], it is necessary to establish four minimum objectives to be followed by the entire global society:

1. Promote employment by creating a sustainable institutional and economic environment
2. Adopting and expanding social protection measures - social security and worker protection - that are sustainable and adapted to national circumstances
3. Promote social dialogue and tripartite as the most appropriate methods
4. Respect, promote, and apply fundamental principles and rights at work, which are of particular importance, both as to rights and as conditions necessary for the full realization of strategic objectives

The working modality mentioned above, click working, is in direct violation of ILO guidelines. In addition to the ILO, the United Nations also provides in Article 23[10] of The Universal Declaration of Human Rights that every individual must receive a dignified, fair, and satisfactory remuneration that assures him/her and his/her family an existence compatible with human dignity. Therefore, it is important to emphasize that, although there are new labor relations, they cannot be allowed in diminishing the value of human labor on behalf of technological advancement.

If the right to decent working and living conditions are universal rights protected by international agencies. And if on a local level such ideas are being sabotaged, shouldn't international agencies have the power to pressure the saboteurs? Perhaps.

Meanwhile, how have the AI Ethics community responded to this possibly inevitable wave of technological unemployment on our way? How can we distribute the new goods and services generated by this economy sustained by intelligent automation? A solution recently proposed by O'Keefe et al. (2020) called the 'Windfall's Clause', a legal *ex-ante*[11] agreement that ensures that companies involved in the AI industry are committed to sharing their profits with society. However, a critique of solutions like the Windfall clause would be that it is just an advertising solution. So that it would appear that there is a plan to combat technological unemployment ('the kindness of the rich and powerful'), and while the wave keeps coming, no real strategy is being implemented.

This 'burden' must be defined democratically and locally so that if the industry acts counterproductively to social goals (e. g., increasing the level of unemployment in a given place), they must be held accountable. While the social ideals to be pursued must be defined locally, the preservation of such a process must be an international concern.

---

[9] http://ilo.org/global/about-the-ilo/mission-and-objectives/WCMS_371208/lang--en/index.htm#:~:text=Adopted%20in%202008%20by%20the,in%20the%20era%20of%20globalization.
[10] http://www.capital.sp.gov.br/cidadao/familia-e-assistencia-social/conheca-seus-direitos/declaracao-universal-dos-direitos-humanos#:~:text=Todo%20ser%20humano%20que%20trabalha,outros%20meios%20de%20prote%C3%A7%C3%A3o%20social.
[11] An agreement is made before the potential event (like large profits from advanced AI development) occurs.

## IV. 'Bring the Rest In!'

It is no surprise that issues such as labor exploitation, violation of humanitarian rights, and mass unemployment, which are often problems related to developing countries, are not raised much in the current debate on AI Ethics. We suggest that this is the result of a debate where most of the published ethical guidelines are produced by a minority of highly developed industrialized countries.

Besides the epistemic values that form our notion of scientific objectivity, like generalization and falsification, there is a strong consensus in the literature that non-epistemic values guide and shape scientific reasoning, and in our case of interest, the interpretation and application of technological developments (Douglas, 2007; Elliott & McKaughan, 2014; Bueter, 2015).

When non-epistemic values guide the technological progress of AI, an obvious question may come to mind: what values are being taken into consideration? When ethical guidelines are written they are composed to reflect the core values of the culture and society responsible for writing them. Not to protect other states, cultures, or segregated populations. But it's the segregated, the periphery of society, that are the first to feel the side effects of the rapid changes caused by our technological advances.

Tools created to optimize processes end up becoming oppressive paraphernalia, favoring certain social groups over others. Examples of how classification algorithms can act in biased ways are not difficult to find. Here we will cite the case of the Brazilian government, which has recently adopted the use of video-monitoring and facial recognition technologies. The Decree Nº 793 [12] of October 2019, proposed by the former Minister of Justice Sérgio Moro, was presented as a way to modernize Brazil's police forces. However, what has been happening is a step backw2ard concerning issues such as transparency, accountability, and protection of personal data.

According to Nunes (2019), the type of policy being adopted has only increased the mass incarceration of segregated populations. First, the facial recognition techniques used by Brazilian police forces are not accurate, something that can generate arbitrary arrests and human rights violations. According to a report made available by the Criminal Defense Coordination and the Board of Studies and Research on Access to Justice of the Public Defender's Office of Rio de Janeiro[13] between June 1, 2019, and March 10, 2020, there were at least 58 cases of false photographic recognition, resulting in unjust accusations, and the imprisonment of innocent individuals. 70% of the unjustly accused were black. Second, Nunes points out that since the implementation of such systems the black population has been disproportionately affected. In 2019, 90.5% of those arrested by facial recognition and video-monitoring systems were black.

Cases of algorithmic bias toward gender and sexual orientation are cited by Costanza-Chock (2018), which showed how intelligent airport screening systems systematically signal transsexual individuals for security searches. A controversial study by Wang and Kosinski (2017), where the authors stated that '*classification algorithms can infer sexual orientation from facial images*', caused a series of criticism from the LGBTQ+ community (Agüera y Arcas et al., 2018). Kosinski made even more controversial statements in an interview for The Guardian (Levin, 2017)

---

[12] Official Journal of the Union - Ordinance No. 793. https://www.in.gov.br/en/web/dou/-/portaria-n-793-de-24-de-outubro-de-2019-223853575.
[13] Public Defender's Office of Rio de Janeiro. http://www.defensoria.rj.def.br/uploads/imagens/d12a8206c9044a3e92716341a99b2f6f.pdf.

# AI Ethics Journal

stating that soon intelligent algorithms will be able to measure IQ, political orientation, and criminal inclinations from *facial images only*. Is the AI community looking to revisit phrenology?

Another unethical aspect is that developing countries are being used as a test area for new technologies. For example, before the involvement with Donald Trump's political campaigns, Ted Cruz, and the separation of the UK from the EU, Cambridge Analytica tested its tools in the 2015 elections of Nigeria and 2017 in Kenya. These countries were chosen for the company's beta phase due to milder data protection laws, which facilitated the unscrupulous use of prediction and classification systems to influence these countries' elections. We see here a clear case of social engineering architected by foreign agents equipped with intelligent autonomous systems (Nyabola, 2018). It is worth noting that the series of lawsuits that led to Cambridge Analytica's closure in 2018 only occurred after its acts against the democracy of countries such as the USA and the UK came to light, and not for its interference on Nigeria or Kenya.

A report entitled 'The Global AI Agenda' written by MIT Technology Review Insights[14] shows that one of the main limitations that Latin American companies reported was the limited participation of Latin America in the development of global governance structures involving the use and development of AI. The European, North American, and Chinese dominance in the making of such guidelines make their integration in the Latin American context difficult, and sometimes impractical.

Carman and Rosman (2020) raised the same issue but focusing on the African continent. The authors argued that the establishment of foreign governance structures is a delicate issue in the African context, given the continent's long history of colonialism and imposition of foreign values.

In 2019 such concerns culminated in several G20 participating countries, such as India, Indonesia, and South Africa, refusing to sign the Osaka Track, an international declaration regulating aspects of e-commerce and data flow from the WTO (World Trade Organization) (Kanth, 2019). The refusal happened because the interests of these countries, as of several others, were not being represented in this document, denying political autonomy for the states themselves to go through their digital industrialization.

In Jobin et al. (2019) and Hagendorff's (2020) meta-analyses, of all reviewed documents, none had a connection with organizations in South America, Africa, or the Middle East. This shows that more than half of the globe is being excluded from the debate about which ethical principles and governance strategies should guide the future transformation of our society. Garcia (2019) points out that virtually the entire Southern Hemisphere is under-represented in the AI governance debate. As most developing countries do not yet have an AI industry capable of competing with more developed countries, the Global South is dependent on the goodwill of other governments in a new colonial technological regime.

For authors like Green (2019), 'good is not good enough', meaning, the limited definitions of what is 'correct' or 'morally justifiable' within areas responsible for technological development, like computer science, software engineering, computer engineering, need to achieve an understanding of what the 'social good' means. And how can we discover what the 'social good' is? Would centralizing the power of choice and regulation in the hands of a few individuals, a technocracy *without elections,* be the best alternative?

---

[14] MIT Technology Review Insights. 2020. The global AI agenda: Latin America. The global AI agenda series. https://mittrinsights.s3.amazonaws.com/AIagenda2020/LatAmAIagenda.pdf.

# AI Ethics Journal

Although still very modest, there is a concern to increase diversity in AI regulation and development. Either by making developing countries and minority groups more active or by seeking to impose new notions of equity, urgency, necessity, and historical restoration in the drafting of norms and guidelines for the tech industry.

For example, ÓhÉigeartaigh et al. (2020) pointed out that Universities have a key role in promoting greater intercultural cooperation on issues related to AI governance and Ethics. Meanwhile, Mohamed et al. (2020) used post-colonial theory to suggest that the needs of marginalized communities, such as developing countries, should be the markers for the creation of governance guidelines for AI.

Citing Humanitarian Rights again, one of the most recent studies to address ethical issues related to notions of human rights in AI governance was the SHERPA survey (Santiago, 2020), commissioned by the European Union. In this study, most of the experts interviewed pointed out that the best strategy to deal with ethical issues involving the use and development of AI is regulation.

Thus, we argue that there is a common theme to be found in these studies:

- The need for regulation
- The need to democratize the governance of AI, making its development more inclusive and democratic

To this end, we would like to draft an agenda composed of three key points to regulate the development of AI. First:

- Create an International Treaty of cooperation, in which a large number of countries should be signatories, including the major superpowers of technological development.

Nowadays, the technological superpowers of the 21st century (United States, China, and the European Union), are in a technological race aimed at the development of advanced AI.

This competition, besides generating a great waste of resources (since everyone is spending time and raw material to achieve the same result), violates principles of solidarity and cooperation, directly damaging our capacity for global coordination.

Instead of acting in an inclusive and cooperative environment, we find ourselves in a context where each agency seeks to impose its values and norms on the world. The details of this treaty are beyond the intent of this study (something of this magnitude could not be defined by two authors, but rather by a democratic process of international deliberation). The only point we emphasize is that the goal of this treaty should be to align the interests of the AI industry with the interests of the general society. Locally, defending the interests expressed by the majority. Globally, by upholding universal notions of human rights and democratic sovereignty.

Second:

- Companies connected to countries that are signatories of the proposed Treaty may be prosecuted (if needed) by an international/impartial agency (e. g., the International Criminal Court or the International Court of Justice) for crimes against humanity.

Thus, we suggest that the precarization of human labor on a massive scale, the automation of processes of oppression against any population, the use of autonomous weapons against human beings, the use of technology as a form of human exploitation, all be treated as crimes against humanity.

# AI Ethics Journal

For this, it would also be necessary to include the right of, e. g., 'Protection against Technological Exploitation and Violence' in the UN Charter of Human Rights or other forms of international humanitarian laws [15]. If international agencies can have jurisdiction over both individuals and companies for crimes against humanity, and since ethical principles do not prevent entire nations from being harmed, perhaps the possibility of a company having to close its doors, and their owners being imprisoned for up to 30 years[16] can be incentive enough.

This form of regulation should not replace local governance. The main goal of international agencies should not be to define what is 'better' or 'right' on a local level. Their aim should be to help the alignment of the AI industry with the society they interact.

For example, in a country like Brazil, where a large part of the workforce is employed through services related to urban mobility, technologies that promote the automation of this process (e. g., autonomous cars) should be critically scrutinized by the public opinion.

Technologies should be created to increase the well-being of the majority of the population. When such ideas aren't being met, international agencies should act (e. g., pressuring local governments to take action).

Furthermore, the Treaty should have an agreement of responsibility and collaboration between companies that seek the development of AI (e. g., Google Brain, DJI, DeepMind, Anki, Open AI), to engage cooperatively towards the development of artificial intelligence. In this way, open-source projects should be the main default for the AI industry.

Democracy is a beautiful idea. However, the concept of 'direct democracy' has been labeled as a mere ideal and an impractical methodology. Recent experiments in more direct and participatory forms of decision making have been tried by, for example, Spain with their 'Decide Madrid' project[17], and Chile whit the 'Senador Virtual'[18]. And what these experiments show is that the main obstacle of direct democracy is not limited communication but the limited cognitive bandwidth of people.

In other words, it is not access to voting that we lack. What we need is something that can 'bust' the cognitive capacities of people to engage in hundreds of different decisions.

Augmented Democracy[19] is the idea of using AI itself to expand the ability of people to participate directly in a high-volume democratic process. There are already small instances of this idea, for example, with 'Sam'[20], a New Zealand virtual politician designed to represent civil individuals collectively. Or with new forms of decentralized learning for preferences elicitation and aggregation, like Federated Learning (Bonawitz et al., 2019).

Perhaps the regulation and governance of the AI industry is the perfect candidate for us to experiment with augmented forms of democracy, to align society's preferences with the objectives of the AI industry.

In the end, like all claims for change, our proposal is not perfect and should be interpreted as a 'criticism' totally passive to be criticized and improved. Our main intention with this article was to instigate the reader to consider the possibility that a new and different form of governance is indeed possible. And who knows, even desirable.

---

[15] International Humanitarian Laws. https://www.icrc.org/en/doc/assets/files/other/what_is_ihl.pdf.
[16] Rome Statute of the International Criminal Court. https://www.icc-cpi.int/resource-library/Documents/RS-Eng.pdf.
[17] A project which empowers citizens of Madrid the ability to vote and decide on public investment. https://decide.madrid.es/.
[18] The project presents citizens with simple questions about some of the bills discussed in the Chilean senate. https://www.senadorvirtual.cl/.
[19] https://www.peopledemocracy.com/.
[20] http://www.politiciansam.nz/.

# AI Ethics Journal

## V. Conclusion

Currently, we see several issues disregarded from the debate on AI Ethics, social issues like income inequality, technological unemployment, humanitarian violations, and the total lack of diversity in the AI and tech industry, should be given a substantial amount of attention. We need to remember that these issues are drastically affecting the lives of millions, if not billions, of individuals today.

Humanitarian rights, the sovereignty of the people, the right to decent labor conditions, and the respect for all diverse expressions of humanity, are far more important problems to be addressed than the privacy of social media users. Online privacy is already a possibility, and the cryptographic community has already shown us the way. Stop using Windows or Mac, learn to use Linux, hide your VPN, use TOR and PGP and be anonymous.

Like critical theory, AI Ethics must focus on highlighting the neglected aspects of our society and its relation to the technological industry, challenging its power structures so that the promise of beneficial AI for all can be fulfilled. Not just as an ideal for the future of humanity, but for the present people too.

And like Cyberpunk, and other forms of anarchical libertarian views, AI governance should be based on a democratic and decentralized form of governance, guaranteeing individual autonomy and freedom for all. At the same time, this right of freedom of choice should be a matter of international, if not universal, care.

# AI Ethics Journal

# AI Ethics Journal

# AI Ethics Journal

# AI Ethics Journal